\documentclass[letterpaper]{article} 
\usepackage{aaai23}  
\usepackage{times}  
\usepackage{helvet}  
\usepackage{courier}  
\usepackage[hyphens]{url}  
\usepackage{graphicx} 
\usepackage{natbib}  
\usepackage{caption} 
\usepackage{algorithm}
\usepackage{algorithmic}

\usepackage{newfloat}
\usepackage{listings}
\DeclareCaptionStyle{ruled}{labelfont=normalfont,labelsep=colon,strut=off} 
\floatstyle{ruled}
\newfloat{listing}{tb}{lst}{}
\floatname{listing}{Listing}
\usepackage{booktabs, makecell} 
\usepackage{verbatim} 
\usepackage{subcaption} 
\usepackage{multirow} 
\usepackage{cite} 
\usepackage{amsmath,amssymb,amsfonts} 

\title{Physics-Based Task Generation through Causal Sequence of Physical Interactions}
\author {
    Chathura Gamage \textsuperscript{\rm 1}, 
    Vimukthini Pinto \textsuperscript{\rm 1}, 
    Matthew Stephenson \textsuperscript{\rm 2}, 
    Jochen Renz \textsuperscript{\rm 1}
}
\affiliations {
    \textsuperscript{\rm 1} School of Computing, The Australian National University, Canberra, Australia\\
    \textsuperscript{\rm 2} College of Science and Engineering, Flinders University, Adelaide, Australia\\
    chathura.gamage@anu.edu.au,
    vimukthini.inguruwattage@anu.edu.au,
    matthew.stephenson@flinders.edu.au,
    jochen.renz@anu.edu.au
}

\usepackage{bibentry}

\begin{document}

\maketitle

\begin{abstract}

Performing tasks in a physical environment is a crucial yet challenging problem for AI systems operating in the real world. Physics simulation-based tasks are often employed to facilitate research that addresses this challenge. In this paper, first, we present a systematic approach for defining a physical scenario using a causal sequence of physical interactions between objects. Then, we propose a methodology for generating tasks in a physics-simulating environment using these defined scenarios as inputs. Our approach enables a better understanding of the granular mechanics required for solving physics-based tasks, thereby facilitating accurate evaluation of AI systems' physical reasoning capabilities. We demonstrate our proposed task generation methodology using the physics-based puzzle game Angry Birds and evaluate the generated tasks using a range of metrics, including physical stability, solvability using intended physical interactions, and accidental solvability using unintended solutions. We believe that the tasks generated using our proposed methodology can facilitate a nuanced evaluation of physical reasoning agents, thus paving the way for the development of agents for more sophisticated real-world applications.

\end{abstract}

\section{Introduction}
Physics-based puzzles are often utilized to assess the physical reasoning abilities of humans \cite{Carmel2000,Cheke2012}, animals \cite{Crows1}, and AI systems \cite{phyq, Phyre}. While humans develop the capacity to perform physical reasoning tasks from their infancy \cite{Valenza2006, permanence}, this has proven to be a challenge for AI systems \cite{phyq, Phyre}. With the expanding use of AI systems in physical environments, there is a need for suitable testbeds to enable the advancement of these systems. Therefore, researchers have developed a range of testbeds in physics-based environments to enable experimentation and evaluation of AI agents' physical reasoning capabilities.

When evaluating the physical reasoning capabilities of an AI system based on its performance on a task, it is essential to have a thorough understanding of the physics mechanics required to solve that task. This enables a rigorous assessment of the AI system's physical reasoning weaknesses based on the physics mechanics necessary to solve the tasks. In some existing benchmarks, this task analysis is not performed at all, or it is done at a high level where the tasks are mainly categorized into families of physical scenarios/events, such as stacking blocks, picking and placing, creating domino effects, physically supporting objects, dropping objects, etc. \cite{ahmed2020causalworld, Allen2020, Physion}. Such manual categorization by developers does not provide a clear understanding of the physics mechanics that distinguish tasks and their associated physical scenarios/events.

Inspired by a widely used approach employed by infant physics researchers in studying physical scenarios \cite{baillargeon2012object, baillargeon2009account, bliss1994force}, this study proposes a method for defining physical scenarios in an environment based on the causal interactions between objects. The definition of physical scenarios is carried out in a granular fashion, considering the causal sequence of physical interactions between objects that are necessary to solve tasks associated with the scenario. The task generation process considers the impact of force and motion on the interactions between objects, establishing the process on the grounds of dynamic physics. This method also paves the way towards a systematic classification of tasks according to the associated physical interactions, filling a gap in the current physical reasoning testbeds and benchmarks.

We have selected the physics-based game Angry Birds as our demonstration domain, as it offers a realistic 2D physical environment and is a popular choice in physical reasoning AI research. To begin, we introduce a grammar that we use to describe tasks and object layouts in physical environments. Using this grammar, we define example physical scenarios as a causal sequence of physical interactions, along with a set of corresponding physical restrictions. We then introduce our task generation process that takes a defined scenario as input and produces a feasible task within our demonstration domain (i.e., an Angry Birds game level). This generation process involves constructing and solving qualitative spatial relationship graphs between objects and satisfying constraints through physics simulations. We evaluate the generated tasks for their physical stability, intended solvability, and accidental solvability, which analyzes whether the tasks are solvable using unintended solutions. Additionally, we analyze the generation time of the task generator as the number of physical interactions in the input physical scenarios increases.

\section{Background and Related Work}
In AI research, the generation of physics-based content is most commonly discussed in physical reasoning testbeds and benchmarks that are used to evaluate AI systems, as well as in physics-based video games when developing game content. In this section, we investigate these two research areas in the context of content generation and position our work in the current literature, emphasizing its contributions.

\subsection{Physical Reasoning Benchmarks and Testbeds}
In recent years, researchers have developed various environments as benchmarks and testbeds to evaluate the physical reasoning capabilities of AI agents. These environments are mainly based on tasks that involve taking actions in a physical environment \cite{phyq, Phyre}, reasoning about images of physical scenarios \cite{hong2021ptr, Cophy}, and reasoning about videos of physical scenarios \cite{IntPhys2019, CLEVRER}. Among them, task generation methods used in action-based environments are related to this work as we also focus on generating tasks that an agent can interact with and take actions to solve them. Examples of recent action-based environments include Phy-Q \cite{phyq}, PHYRE \cite{Phyre}, Virtual Tools \cite{Allen2020}, OGRE \cite{OGRE}, CausalWorld \cite{ahmed2020causalworld}, and RLBench \cite{james2020rlbench}.

The design and development of tasks in these works are predominantly done manually by developers, with most utilizing automated generation techniques to create simple variations of handcrafted task templates.  For example, Phy-Q testbed has handcrafted task templates for 15 physical scenarios, and tasks are generated from those templates by slightly varying the locations of objects in the template and adding distracting objects. 
PHYRE, Virtual Tools, and OGRE follow a similar procedure. They all have pre-designed task templates and tasks are generated by either varying the shape, size, and location of the original template. 
CausalWorld is a robotic benchmark with eight types of tasks, such as pushing, picking, picking and placing, and stacking.
They have separate generators to generate tasks for each of those task types. The generators have hard-coded templates, and the tasks are generated by sampling new tasks from those templates.

Despite the use of automated generation techniques to some extent in these benchmarks and testbeds, task generation is still heavily reliant on pre-defined templates. The process of creating these templates can be tedious and time-consuming, as the template developer must ensure that the template is stable under gravity and other physical constraints, solvable, and flexible enough to allow for modifications in task generation. 
In contrast, our proposed generation method does not require any initial task templates. Instead, the input to the generator is a minimal description of the physical scenario, defined as a sequence of physical interactions between objects that leads to solving the task. This approach eliminates the need for pre-designed templates and simplifies the task generation. 
Furthermore, tasks generated from this method consider the physical interactions involved in solving the task, allowing for systematic categorization of tasks into distinct physical scenarios based on the involved physical interactions. This feature is particularly valuable for physical reasoning testbeds and benchmarks, as it can be used to classify tasks systematically and evaluate AI agents' strengths and weaknesses more comprehensively. The current task classifications on those benchmarks do not justify why a given task belongs to its class.

\subsection{Physics Based Puzzle Games}

Physics-based puzzle games have gained popularity in the field of Procedural Content Generation (PCG) owing to the intriguing physics-related challenges that are encountered while generating content for such games, which are applicable to real-world physics challenges. Popular physics puzzle games, such as Angry Birds \cite{2017AIBirdsComp} and Cut the Rope \cite{cutTheRope}, have been used as test domains by researchers to investigate physics-based content generation techniques. Of these games, Angry Birds has received considerable attention in the research community, particularly in the fields of PCG \cite{2017AIBirdsComp} and AI game-playing agent development \cite{Renzet2019}. Consequently, in this study, we utilize Angry Birds as a test domain to showcase our proposed task generation methodology.

\begin{figure}
    \centering
    \includegraphics[width=1.0\linewidth]{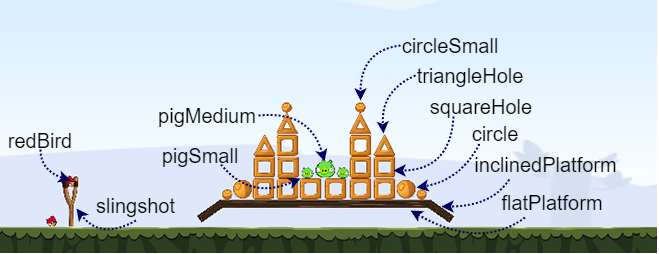}
    \caption{A game level from Angry Birds.}
    \label{example-AB-game-level}
\end{figure}

Angry Birds is a 2D physics simulation game wherein players must destroy pigs in a given game level by launching a designated number of birds from a slingshot. The levels in the game consist of dynamic objects (birds, blocks, and pigs) that adhere to Newtonian physics and static objects (platforms) that are not influenced by external forces. The dynamic objects possess health points that decrease upon collision and are destroyed when their health points are completely depleted.
In this study, we use Science Birds \cite{scienceBirds}, a research clone of the game developed in Unity with the Box2D physics engine. A modified version of Science Birds, with adjusted physics parameters and health points of objects, is employed to effectively showcase the proposed methodology. The modifications address limitations in Science Birds' object dynamics and fragility, ensuring a more accurate demonstration of the methodology.
Figure \ref{example-AB-game-level} displays a game level from Angry Birds in Science Birds, showcasing the primary game objects utilized in this study.

Numerous research studies have explored various methods for creating game levels in Angry Birds. 
Prior investigations have tackled this problem from diverse angles, proposing solutions to intriguing challenges, such as generating stable structures within the physical environment \cite{stephenson2017generating}, dynamically adjusting the game levels' difficulty \cite{stephenson2019agent}, creating block structures based on hand-drawn sketches \cite{sketchesToLevels}, generating deceptive levels to deceive AI agents \cite{deceptiveGenerator}, generating novel scenarios for physics environments \cite{gamage2023novphy, gamage2021novelty} and, most recently, using prompt engineering to create prompts for level generation \cite{taveekitworachai2023chatgpt4pcg}.
None of the previous studies has treated the level generation for Angry Birds as a task generation problem that incorporates the physical interactions between the game objects. This paper proposes a novel approach that defines a task as a causal sequence of physical interactions and generates tasks whose solution follows this sequence. The generation process considers the effect of force and motion on the interactions of objects, resulting in a more scientifically grounded approach in terms of dynamic physics. By generating tasks based on the physical interactions present in the solution, agents can be evaluated and compared based on their performance in specific interaction sequences of interest. Therefore, the proposed generation method enables a rigorous evaluation of AI agents' physical reasoning abilities, similar to the evaluation conducted in the Phy-Q testbed that uses handcrafted Angry Birds levels.

\section{Grammar and Defining Scenarios}

This section presents the proposed grammar and discusses how the grammar can be used to define physical scenarios.

\subsection{Grammar for Angry Birds}

The introduced grammar is used to describe objects, interactions, restrictions, and object layouts in a physical environment. The grammar consists of four components, namely, object grammar, interaction grammar, restriction grammar, and layout grammar. The first three grammar components are utilized to define physical scenarios, while the last grammar component is utilized in the generation process for defining the object layouts within the physical environment.

\subsubsection{Object Grammar}
The Object grammar component is used to define various types of objects that may exist in a physical environment. Each grammar term represents a class of objects that share similar properties, such as the ability to roll or slide. These terms are listed in Table \ref{object-grammar}. In Angry Birds, the objects include birds that the player shoots (a single type is used in this work, named redBird), pigs that the player has to destroy (two types with different sizes are used, named pigSmall and pigMedium), blocks with different shapes and sizes (including circle, circleSmall, squareHole, and triangleHole), and platform objects that can vary in size and rotation (mainly used as surfaces).

\begin{table}
  \scriptsize
  \centering
  {\fontsize{9}{11}\selectfont
  \begin{tabular}{ll}
    \toprule
     Grammar Term & Game Objects Represented \\
    \midrule
    bird & redBird \\
    pig & pigSmall$\vee$pigMedium \\
    rollableBlock & circleSmall$\vee$circle \\
    fallableBlock & circleSmall$\vee$circle$\vee$squareHole$\vee$\\
    & triangleHole\\
    slidableBlock & squareHole$\vee$triangleHole \\
    horizontalSurface & flatPlatform \\
    inclinedSurface & inclinedPlatform \\
    surface & flatPlatform$\vee$inclinedPlatform\\
    \bottomrule
  \end{tabular}}
  \caption{Object grammar.}
  \label{object-grammar}
\end{table}

\subsubsection{Interaction Grammar} 
Interaction grammar defines fundamental physical interactions between objects. When defining the interaction terms, we take into account the impact of force and motion, drawing inspiration from experiments that were conducted to understand the physical reasoning capabilities of infants \cite{bliss1994force}.  
Specifically, we focus on the fundamental physical interaction where one object applies a force on another object; in Angry Birds, which commonly occurs by one object hitting another object. We then analyze the consequent effects of the force, which can lead to the dynamics of the object being affected, causing it to roll, fall, slide, or bounce, or even making the object deform or get destroyed as seen in Angry Birds. The interaction grammar terms are listed in Table \ref{interaction-grammar}.

\begin{table}
  \scriptsize
  \centering
  {\fontsize{9}{11}\selectfont
  \begin{tabular}{ll}
    \toprule
     Grammar Term & Description\\
    \midrule
    hit(\textit{a})(\textit{b})(\textit{d}) & \textit{a} collides with \textit{b} from direction \textit{d} of \textit{b} \\ 
    \noalign{\vspace{-0.7ex}}
    & \textit{d} $\in$ \{\textit{left, right, above, below, any}\} \\
    roll(\textit{a})(\textit{b})(\textit{d}) & \textit{a} rolls on \textit{b} towards direction \textit{d}\\ 
    \noalign{\vspace{-0.7ex}}
    & \textit{d} $\in$ \{\textit{left, right}\} \\
    fall(\textit{a})(\textit{b}) & \textit{a} falls towards \textit{b}\\
    slide(\textit{a})(\textit{b})(\textit{d}) & \textit{a} slides on \textit{b} towards direction \textit{d}\\ 
    \noalign{\vspace{-0.7ex}}
    & \textit{d} $\in$ \{\textit{left, right}\} \\
    bounce(\textit{a})(\textit{b})(\textit{d}) & \textit{a} bounces off \textit{b} towards direction \textit{d}\\ 
    \noalign{\vspace{-0.7ex}}
    & \textit{d} $\in$ \{\textit{left, right, above, below}\} \\
    destroy(\textit{a})(\textit{b}) & \textit{a} destroys \textit{b} in the collision with \textit{b}\\
    \bottomrule
  \end{tabular}}
  \caption{Interaction grammar. The parameters \textit{a} and \textit{b} represent objects, and \textit{d} represents a direction. }
  \label{interaction-grammar}
\end{table}

\subsubsection{Restriction Grammar} 
The restriction grammar defines the restricted interactions between objects. Since an object in a physical environment has many possibilities of interaction, restriction grammar narrows down the interactions of interest of an object. Our current restriction terms, as shown in Table \ref{restriction-grammar}, prevent objects from hitting each other and from falling in their movements. These restrictions are put in place as hitting is the primary way of force transferring in Angry Birds, and falling is a natural phenomenon that occurs when an object is placed in the physical environment. It is essential to note that these restrictions only apply to the solution interaction sequence of the task and not to any possible interaction of the relevant object. For instance, an object with a restriction of cannotFall in a task implies that the object cannot fall when the solution interactions of the task are being executed, but not that it cannot fall by any means when someone is interacting with the environment.

\begin{table}[t]
  \small
  \scriptsize
  \centering
  {\fontsize{9}{11}\selectfont
  \begin{tabular}{ll}
    \toprule
     Grammar Term & Description\\
    \midrule
    cannotHit(\textit{a})(\textit{b})(\textit{d}) & \textit{a} cannot collide with \textit{b} from direction \textit{d}\\ 
    \noalign{\vspace{-0.7ex}}
    &  of \textit{b}, $\textit{d} \in \{\textit{left, right, above, below, any}\}$ \\ 
    cannotFall(\textit{a}) & \textit{a} cannot fall in its motion\\
    \bottomrule
  \end{tabular}}
  \caption{Restriction grammar. The parameters \textit{a} and \textit{b} represent objects, and \textit{d} represents a direction.}
  \label{restriction-grammar}
\end{table}

\subsubsection{Layout Grammar} 
\label{layout-grammar-section}
Layout grammar encompasses a set of terms that characterize spatial relationships between objects in a physical environment. This grammar is utilized by the generator to specify the configuration of objects during task generation. The layout grammar terms deemed appropriate for defining spatial relationships within the context of Angry Birds are displayed in Table \ref{layout-grammar}.

\begin{table}[t]
  \scriptsize
  \centering
  {\fontsize{9}{11}\selectfont
  \begin{tabular}{ll}
    \toprule
     Grammar Term & Description\\
    \midrule
    inDirection(\textit{a})(\textit{b})(\textit{d}) & \textit{a} is in direction \textit{d} of \textit{b}\\ \noalign{\vspace{-0.7ex}}
    & \textit{d} $\in$ \{\textit{left, right, above, below}\} \\
    onLocation(\textit{a})(\textit{b})(\textit{l}) & \textit{a} is on top of \textit{b} at location \textit{l}\\ 
    \noalign{\vspace{-0.7ex}}
    & \textit{l} $\in$ \textit{L}, \textit{L} = \{\textit{left, centre, right}\} \\
    locatedFar(\textit{a})(\textit{b})(\textit{d}) & \textit{a} is far from \textit{b} in direction d of \textit{b}\\ 
    \noalign{\vspace{-0.7ex}}
    & \textit{d} $\in$ \{\textit{left, right, above, below}\} \\
    touching(\textit{a})(\textit{b})(l) & \textit{a} touches \textit{b} at location \textit{l} of \textit{b}, \textit{l} $\in$\\ 
    \noalign{\vspace{-0.7ex}}
    & \{\textit{upperLeft, centreLeft, lowerLeft}\} \\
    pathObstructed(\textit{a})(\textit{b})(\textit{d}) & there is an obstacle in the path from\\ 
    \noalign{\vspace{-0.7ex}}
    & \textit{a} and \textit{b}, in the direction \textit{d} to \textit{b}, \\
    \noalign{\vspace{-0.7ex}}
    &\textit{d} $\in$ \{\textit{left, right, above, below, all}\} \\
    liesOnPath(\textit{a})(\textit{b}) & \textit{a} lies on \textit{b}’s moving path\\ 
    \bottomrule
  \end{tabular}}
  \caption{Layout grammar. The parameters \textit{a} and \textit{b} represent objects, \textit{d} represents a direction, and \textit{l} represents a location.}
  \label{layout-grammar}
\end{table}

\subsection{Defining Scenarios}

\begin{table*}[!ht]
  \scriptsize
  \centering
  {\fontsize{9}{11}\selectfont
  \begin{tabular}{ll}
    \toprule
     Name & Scenario Definition\\
     
    \midrule
    1. SF & \{[hit(\textit{bird})(\textit{pig})(\textit{any})] $>$ [destroy(\textit{bird})(\textit{pig})]\}, \{\} \\ 
    \midrule
    2. SFTB & \{[hit(\textit{bird})(\textit{pig})(\textit{any})] $>$ [destroy(\textit{bird})(\textit{pig})]\}, \{[cannotHit(\textit{bird})(\textit{pig})(\textit{above})]\} \\ 
    \midrule
    3. SFLB & \{[hit(\textit{bird})(\textit{pig})(\textit{any})] $>$ [destroy(\textit{bird})(\textit{pig})]\}, \{[cannotHit(\textit{bird})(\textit{pig})(\textit{left})]\} \\ 
    
    \midrule
    4. R & \{[hit(\textit{bird})(\textit{rBlock})(\textit{left})] $>$ [roll(\textit{rBlock})(\textit{surface})(\textit{right})] $>$ [hit(\textit{rBlock})(\textit{pig})(\textit{left})] $>$ [destroy(\textit{rBlock})(\textit{pig})]\}, \\ & \{[cannotHit(\textit{bird})(\textit{pig})(\textit{any})]$\wedge$[cannotFall(\textit{rBlock})]\} \\ 
    \midrule
    5. F & \{[hit(\textit{bird})(\textit{fBlock})(\textit{left}$\vee$\textit{above})] $>$ [fall(\textit{fBlock})(\textit{pig})] $>$ [hit(\textit{fBlock})(\textit{pig})(\textit{above})] $>$ [destroy(\textit{fBlock})(\textit{pig})]\},  \\ & \{[cannotHit(\textit{bird})(\textit{pig})(\textit{any})]\} \\ 
    \midrule
    6. S & \{[hit(\textit{bird})(\textit{sBlock})(\textit{left})] $>$ [slide(\textit{sBlock})(\textit{hSurface})(\textit{right})] $>$ [hit(\textit{sBlock})(\textit{pig})(\textit{left})] $>$ [destroy(\textit{sBlock})(\textit{pig})]\}, \\ & \{[cannotHit(\textit{bird})(\textit{pig})(\textit{any})]$\wedge$[cannotFall(\textit{sBlock})]\} \\ 
    \midrule
    7. B & \{[hit(\textit{bird})(\textit{iSurface})(\textit{any})] $>$ [bounce(\textit{bird})(\textit{iSurface})(\textit{below})] $>$ [hit(\textit{bird})(\textit{pig})(\textit{above})] $>$ [destroy(\textit{bird})(\textit{pig})]\},  \\ & \{[cannotHit(\textit{bird})(\textit{pig})(\textit{any})]\} \\ 
    
    \midrule
    8. RF & \{[hit(\textit{bird})(\textit{rBlock})(\textit{left})] $>$ [roll(\textit{rBlock})(\textit{surface})(\textit{right})] $>$ [hit(\textit{rBlock})(\textit{fBlock})(\textit{left})] $>$ [fall(\textit{fBlock})(\textit{pig})] $>$ \\ & [hit(\textit{fBlock})(\textit{pig})(\textit{above})] $>$ [destroy(\textit{fBlock})(\textit{pig})]\}, \{[cannotHit(\textit{bird})(\textit{pig})(\textit{any})]$\wedge$[cannotFall(\textit{rBlock})]\} \\ 
    \midrule
    9. RS & \{[hit(\textit{bird})(\textit{rBlock})(\textit{left})] $>$ [roll(\textit{rBlock})(\textit{surface})(\textit{right})] $>$ [hit(\textit{rBlock})(\textit{sBlock})(\textit{left})] $>$ [slide(\textit{sBlock})(\textit{hSurface})(\textit{right})] $>$ \\ & [hit(\textit{sBlock})(\textit{pig})(\textit{left})] $>$ [destroy(\textit{sBlock})(\textit{pig})]\}, \{[cannotHit(\textit{bird})(\textit{pig})(\textit{any})]$\wedge$[cannotFall(\textit{rBlock})]$\wedge$[cannotFall(\textit{sBlock})]\} \\ 
    \midrule
    10. FR & \{[hit(\textit{bird})(\textit{fBlock})(\textit{left}$\vee$\textit{above})] $>$ [fall(\textit{fBlock})(\textit{rBlock})] $>$ [hit(\textit{fBlock})(\textit{rBlock})(\textit{above}$\vee$\textit{left})] $>$ [roll(\textit{rBlock})(\textit{surface})\\ &(\textit{right})] $>$ [hit(\textit{rBlock})(\textit{pig})(\textit{left})] $>$ [destroy(\textit{rBlock})(\textit{pig})]\}, \{[cannotHit(\textit{bird})(\textit{pig})(\textit{any})]$\wedge$[cannotFall(\textit{rBlock})]\} \\ 
    \midrule
    11. SR & \{[hit(\textit{bird})(\textit{sBlock})(\textit{left})] $>$ [slide(\textit{sBlock})(\textit{hSurface})(\textit{right})] $>$ [hit(\textit{sBlock})(\textit{rBlock})(\textit{left})] $>$ [roll(\textit{rBlock})(\textit{surface})(\textit{right})] $>$ \\ & [hit(\textit{rBlock})(\textit{pig})(\textit{left})] $>$ [destroy(\textit{rBlock})(\textit{pig})]\}, \{[cannotHit(\textit{bird})(\textit{pig})(\textit{any})]$\wedge$[cannotFall(\textit{sBlock})]$\wedge$[cannotFall(\textit{rBlock})]\} \\ 
    \midrule
    12. BF & \{[hit(\textit{bird})(\textit{iSurface})(\textit{any})] $>$ [bounce(\textit{bird})(\textit{iSurface})(\textit{below})] $>$ [hit(\textit{bird})(\textit{fBlock})(\textit{left}$\vee$\textit{above})] $>$ \\ & [fall(\textit{fBlock})(\textit{pig})] $>$ [hit(\textit{fBlock})(\textit{pig})(\textit{above})] $>$ [destroy(\textit{fBlock})(\textit{pig})]\}, \{[cannotHit(\textit{bird})(\textit{pig})(\textit{any})]\} \\ 

    \midrule
    13. SRF & \{[hit(\textit{bird})(\textit{sBlock})(\textit{left})] $>$ [slide(\textit{sBlock})(\textit{hSurface})(\textit{right})] $>$ [hit(\textit{sBlock})(\textit{rBlock})(\textit{left})] $>$ [roll(\textit{rBlock})(\textit{surface})(\textit{right})] $>$ \\ & [hit(\textit{rBlock})(\textit{fBlock})(\textit{left})] $>$ [fall(\textit{fBlock})(\textit{pig})] $>$  [hit(\textit{fBlock})(\textit{pig})(\textit{any})] $>$ [destroy(\textit{fBlock})(\textit{pig})]\}, \\ & \{[cannotHit(\textit{bird})(\textit{pig})(\textit{any})]$\wedge$[cannotFall(\textit{sBlock})]$\wedge$[cannotFall(\textit{rBlock})]\} \\
    \midrule
    14. SFR & \{[hit(\textit{bird})(\textit{sBlock})(\textit{left})] $>$ [slide(\textit{sBlock})(\textit{hSurface})(\textit{right})] $>$ [hit(\textit{sBlock})(\textit{fBlock})(\textit{left})] $>$ [fall(\textit{fBlock})(\textit{rBlock})] $>$ \\ & [hit(\textit{fBlock})(\textit{rBlock})(\textit{left}$\vee$\textit{above})] $>$ [roll(\textit{rBlock})(\textit{surface})(\textit{right})] $>$ [hit(\textit{rBlock})(\textit{pig})(\textit{any})] $>$ [destroy(\textit{rBlock})(\textit{pig})]\}, \\ & \{[cannotHit(\textit{bird})(\textit{pig})(\textit{any})]$\wedge$[cannotFall(\textit{sBlock})]$\wedge$[cannotFall(\textit{rBlock})]\} \\
    \midrule
    15. RRF & \{[hit(\textit{bird})(\textit{rBlock1})(\textit{left})] $>$ [roll(\textit{rBlock1})(surface1)(\textit{right})] $>$ [hit(\textit{rBlock1})(\textit{rBlock2})(\textit{left})] $>$ \\ &[roll(\textit{rBlock2})(surface2)(\textit{right})] $>$ [hit(\textit{rBlock2})(\textit{fBlock})(\textit{left})] $>$ [fall(\textit{fBlock})(\textit{pig})] $>$ [hit(\textit{fBlock})(\textit{pig})(\textit{any})] $>$ \\ & [destroy(\textit{fBlock})(\textit{pig})]\}, \{[cannotHit(\textit{bird})(\textit{pig})(\textit{any})]$\wedge$[cannotFall(\textit{rBlock1})]$\wedge$[cannotFall(\textit{rBlock2})]\} \\
    \midrule
    16. RRR & \{[hit(\textit{bird})(\textit{rBlock1})(\textit{left})] $>$ [roll(\textit{rBlock1})(surface1)(\textit{right})] $>$ [hit(\textit{rBlock1})(\textit{rBlock2})(\textit{left})] $>$  [roll(\textit{rBlock2})(surface2) \\ & (\textit{right})] $>$ [hit(\textit{rBlock2})(\textit{rBlock3})(\textit{left})] $>$ [roll(\textit{rBlock3})(surface3)(\textit{right})] $>$ [hit(\textit{rBlock3})(\textit{pig})(\textit{any})] $>$ [destroy(\textit{rBlock3})\\ &(\textit{pig})]\}, \{[cannotHit(\textit{bird})(\textit{pig})(\textit{any})]$\wedge$[cannotFall(\textit{rBlock1})]$\wedge$[cannotFall(\textit{rBlock2})]$\wedge$[cannotFall(\textit{rBlock3})]\} \\
    \bottomrule
  \end{tabular}}
  \caption{Definitions of 16 example physical scenarios. In the definition, a sequence of physical interactions (order denoted by $>$) is followed by a set of restrictions. The abbreviations in the names of scenarios 1-7 are SF (Single Force), SFTB (Single Force Top Blocked), SFLB (Single Force Left Blocked), R (Rolling), F (Falling), S (Sliding), and B (Bouncing). The remaining scenario names are formed by combining R, F, S, and B (e.g., RF represents Rolling Falling and SRF represents Sliding Rolling Falling). The object grammar terms rollableBlock, fallableBlock, slidableBlock, horizontalSurface, and inclinedSurface are abbreviated as rBlock, fBlock, sBlock, hSurface, and iSurface. 
  For overloaded parameter values, any of the overloaded values can be used (e.g., hit(\textit{bird})(\textit{fBlock})(\textit{left}$\vee$\textit{above}) represents the bird collides with the \textit{fBlock} from \textit{left} or from \textit{above}).
  }
  \label{example-scenario-definitions}
\end{table*}

As stated previously, the object grammar, interaction grammar, and restriction grammar components are used to define the scenarios. The initial step of defining a scenario involves identifying the objects that are relevant to the scenario, which are characterized using the object grammar. The next step is to determine the intended sequence of physical interactions that must take place between these objects to obtain a solution for the scenario. These interactions are described using the interaction grammar and are arranged sequentially based on their causality. If any interaction needs to be restricted between the objects in the scenario, the restriction grammar is employed to add them.

Table \ref{example-scenario-definitions} displays 16 sample scenarios, which were defined for illustrative purposes in this study. Scenarios one through seven were created to replicate the physical scenarios of Single Force (1 to 3), Rolling (4), Falling (5), Sliding (6), and Bouncing (7) in the Phy-Q testbed. For instance, in the first scenario, the bird collides with the pig from any direction, causing the pig to be destroyed, and there are no associated restrictions. In the second and third scenarios, the bird cannot collide with the pig from above or from the left, respectively, due to imposed restrictions. In the fourth scenario, the bird hits a rollable block, causing it to roll on a surface and then collide with a pig, leading to its destruction. The restrictions in this scenario state that the bird cannot collide directly with the pig (i.e., the player cannot shoot the bird directly at the pig to solve the task) and that the rollable block must not fall during its movement. Scenarios eight to 12 combine two out of Rolling, Falling, Sliding, and Bouncing, while scenarios 13 to 16 combine three of them.

\section{Task Generation Process}

The proposed methodology for generating tasks is discussed in detail in this section as a six-step process. Firstly, the layout constraints between objects are inferred and represented using a layout constraint graph based on the task definition. Secondly, Qualitative Spatial Relations (QSRs) between objects are inferred and the layout constraint graph is converted into a QSR graph. To verify consistency and solve constraints to obtain the initial positions of the objects, the spatial constraints are projected into X and Y dimensions separately and dimension graphs are generated and solved in the third step. Next, non-QSRs of the objects, such as restrictions related to object destruction, are satisfied using simulations and the final positions of the objects are determined. In the next step, distractions are added to the tasks, and finally, the solvability of the tasks is verified.

\subsection{Generating the Layout Constraint Graph}
\label{abc}

\begin{table}[!h]
  \small
  \scriptsize
  \centering
  {\fontsize{9}{11}\selectfont
  \begin{tabular}{ll}
    \toprule
    Predicate & Inferred Layout Constraints \\
    \midrule
    hit(\textit{a})(\textit{b})(\textit{d})& liesOnPath(\textit{b})(\textit{a})$\wedge$inDirection(\textit{b})(\textit{a})(\textit{-d}) \\
    roll(\textit{a})(\textit{b})(\textit{d}) & inDirection(\textit{a})(\textit{b})(\textit{-d})\\
    fall(\textit{a})(\textit{b})(\textit{d}) & locatedFar(\textit{a})(\textit{b})(\textit{d})\\
    slide(\textit{a})(\textit{b})(\textit{d}) & inDirection(\textit{a})(\textit{b})(\textit{-d})\\
    bounce(\textit{a})(\textit{b})(\textit{d}) & inDirection(\textit{a})(\textit{b})(\textit{d})\\
    cannotHit(\textit{a})(\textit{b})(\textit{d}) & pathObstructed(\textit{a})(\textit{b})(\textit{d})\\ 
    cannotFall(\textit{a}) & touching(\textit{a})(\textit{p})(\textit{l\textsubscript{}p})$\wedge$touching(\textit{p})(\textit{q})(\textit{l\textsubscript{}q}),... \\
    \bottomrule
  \end{tabular}}
  \caption{Inferred layout constraints from the interactions and restrictions. The parameter \textit{-d} refers to the opposite direction of \textit{d}. In the term cannotFall(\textit{a}), \textit{p}, \textit{q}, and so on denote the objects that \textit{a} moves on top of in the specified order, while \textit{l\textsubscript{}p}, \textit{l\textsubscript{}q}, and so on signify the locations where those objects are connected, forming a continuous path for \textit{a} to move along.}
  \label{inferring-layout-constraints}
\end{table}

\begin{figure}[!h]
    \centering
    \begin{subfigure}{0.34\textwidth}
        \includegraphics[width=\textwidth]{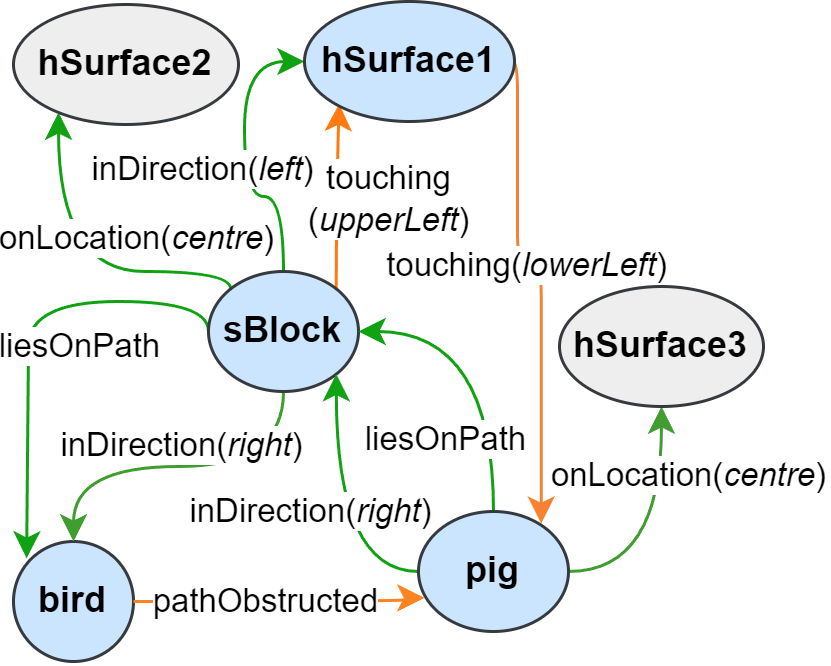}
        \caption{6. S (Sliding) scenario.}
        \label{example-layout-graphs:subfig1}
    \end{subfigure}
    \hfill
    \begin{subfigure}{0.37\textwidth}
        \includegraphics[width=\textwidth]{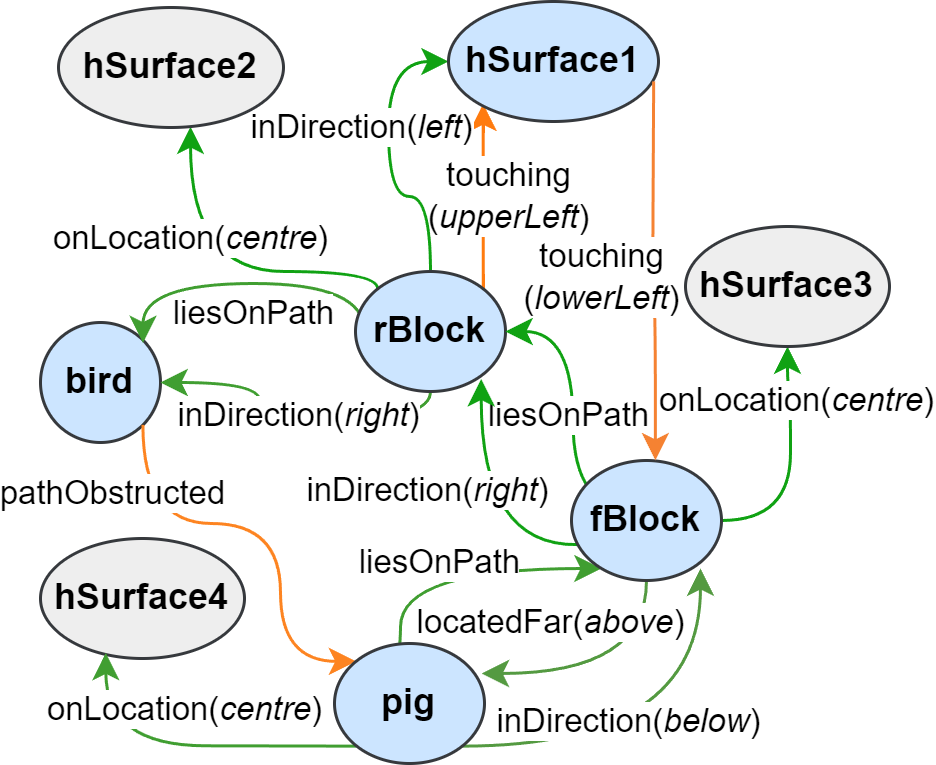}
        \caption{9. RF (Rolling Falling) scenario.}
        \label{fig:subfig2}
    \end{subfigure}
    \caption{
    The layout constraint graphs for scenarios 6 and 9. Blue nodes represent objects in the scenario definition, while grey nodes are the objects introduced to ensure stability under gravity. Green edges represent layout constraints inferred from interactions, while orange edges represent those inferred from restricted interactions.}
    \label{example-layout-graphs}
\end{figure}

The generation process begins by determining the objects required to be present in the scenario and inferring the layout constraints between the objects. To achieve this, the scenario definition is used to identify all objects required for the scenario. Then, the layout constraints are inferred based on the interactions and restricted interactions between objects and the constraints are described using the layout grammar. The layout constraints that can be inferred are shown in Table \ref{inferring-layout-constraints}. The layout constraint graph is then generated with the objects in the scenario as nodes and the directed edges representing the layout constraints between the objects. In this process, a layout optimization is conducted to remove redundant layout constraints and ensure physical stability. This stability enhancement involves confirming that dynamic objects are supported by static ones and adding static objects beneath those not adequately supported, ensuring equilibrium under gravity assuming that the task is initially stable. Figure \ref{example-layout-graphs} presents the layout graphs of scenarios 6 and 9.

\subsection{Transforming Layout Constraints into Spatial Relations}

\begin{figure}[!h]
    \centering
    \includegraphics[width=0.473\textwidth]{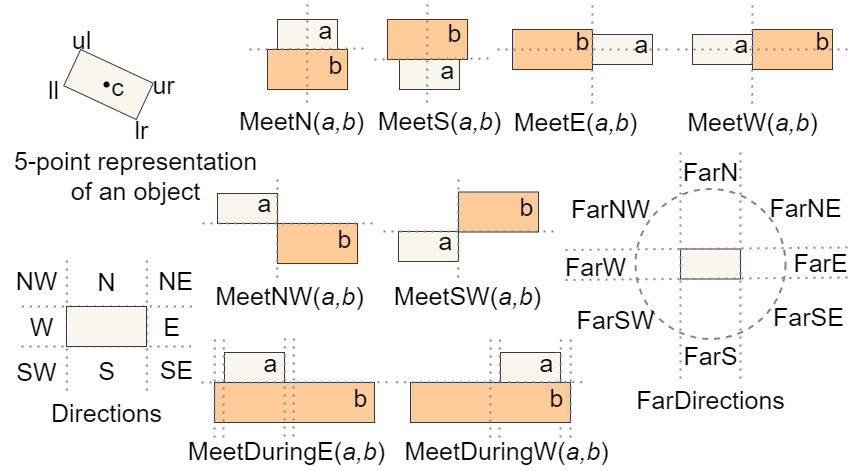}
    \caption{Illustrations of qualitative spatial relations.}
    \label{qsr-relationships-illustration}
\end{figure}

In this phase, the graph of layout constraints is transformed into a spatial relationship graph among objects in 2D Euclidean space. To accomplish this, we draw inspiration from various QSR calculi, including cardinal direction relations \cite{cardinalDirectionsQSR}, interval relationships \cite{allen1983maintaining}, and topological relationships \cite{topologicalRelations}. To meet the requirements of this study, we introduce a range of spatial predicates that are relevant to objects in 2D Euclidean space. Usually, in QSR literature, 2D objects are represented using their Minimum Bounding Rectangle (MBR) \cite{QSRsurvey}. However, this approach has limitations when determining the actual corner positions of an object if it is rotated, which is important for determining the precise placement of neighbouring objects, especially when satisfying the layout terms such as touching. To overcome this, we suggest a 5-point representation for a 2D object using the positions of 5 points lower left (ll), centre (c), upper right (ur), upper left (ul), and lower right (lr). When defining QSRs, we assume that objects have a maximum of 90 degrees of rotation. The defined QSRs are illustrated in Figure \ref{qsr-relationships-illustration}. The layout constraints are translated into spatial constraints using these QSRs. The QSRs in our defined set that can be inferred for the layout constraints are shown in Table \ref{inferred-qsr-relations}. 

As can be seen from Table \ref{inferred-qsr-relations}, a single layout constraint can be mapped to various QSRs. Additionally, the generator can leverage the flexibility in defining the scenario by overloading the parameters of the grammar terms, such as in scenario 10, allowing the bird to hit the fBlock from \textit{either the left or the above} and enabling the fBlock to hit the rBlock from \textit{either the above or the left}. These opportunities for flexibility in the generation process enable the generation of a varied range of layout configurations for a given scenario definition.

\begin{table}[t]
  \scriptsize
  \centering
  {\fontsize{9}{11}\selectfont
  \begin{tabular}{ll}
    \toprule
     Layout Predicate & Inferred QSRs\\
    \midrule
    inDirection(\textit{a})(\textit{b})(\textit{left}) & W(\textit{a,b})$\vee$NW(\textit{a,b})$\vee$SW(\textit{a,b})\\
    inDirection(\textit{a})(\textit{b})(\textit{right}) & E(\textit{a,b})$\vee$NE(\textit{a,b})$\vee$SE(\textit{a,b})\\
    inDirection(\textit{a})(\textit{b})(\textit{above}) & N(\textit{a,b})$\vee$NE(\textit{a,b})$\vee$NW(\textit{a,b})\\
    inDirection(\textit{a})(\textit{b})(\textit{below}) & S(\textit{a,b})$\vee$SE(\textit{a,b})$\vee$SW(\textit{a,b})\\
    \hline
    onLocation(\textit{a})(\textit{b})(\textit{left}) & MeetDuringW(\textit{a,b})\\
    onLocation(\textit{a})(\textit{b})(\textit{centre}) & MeetN(\textit{a,b})\\
    onLocation(\textit{a})(\textit{b})(\textit{right}) & MeetDuringE(\textit{a,b})\\
    \hline
    locatedFar(\textit{a})(\textit{b})(\textit{left}) & FarW(\textit{a,b})$\vee$FarNW(\textit{a,b})$\vee$ \\ \noalign{\vspace{-0.7ex}} &
    FarSW(\textit{a,b})\\
    locatedFar(\textit{a})(\textit{b})(\textit{right}) & FarE(\textit{a,b})$\vee$FarNE(\textit{a,b})$\vee$ \\ \noalign{\vspace{-0.7ex}} & FarSE(\textit{a,b})\\
    locatedFar(\textit{a})(\textit{b})(\textit{above}) & FarN(\textit{a,b})$\vee$FarNE(\textit{a,b})$\vee$ \\ \noalign{\vspace{-0.7ex}} & FarNW(\textit{a,b})\\
    locatedFar(\textit{a})(\textit{b})(\textit{below}) & FarS(\textit{a,b})$\vee$FarSE(\textit{a,b})$\vee$ \\ \noalign{\vspace{-0.7ex}} & FarSW(\textit{a,b})\\
    \hline
    touching(\textit{a})(\textit{b})(\textit{upperLeft}) & MeetNW(\textit{a,b})\\
    touching(\textit{a})(\textit{b})(\textit{centreLeft}) & MeetW(\textit{a,b})\\
    touching(\textit{a})(\textit{b})(\textit{lowerLeft}) & MeetSW(\textit{a,b})\\
    \bottomrule
  \end{tabular}}
  \caption{Inferred QSRs from the layout constraints. The abbreviated notation of the directions signifies their conventional representations.}
  \label{inferred-qsr-relations}
\end{table}

\subsection{Generating a Plausible Spatial Configuration for QSR Constraints}

\begin{figure}[t]
    \centering
    \includegraphics[width=0.46\textwidth]{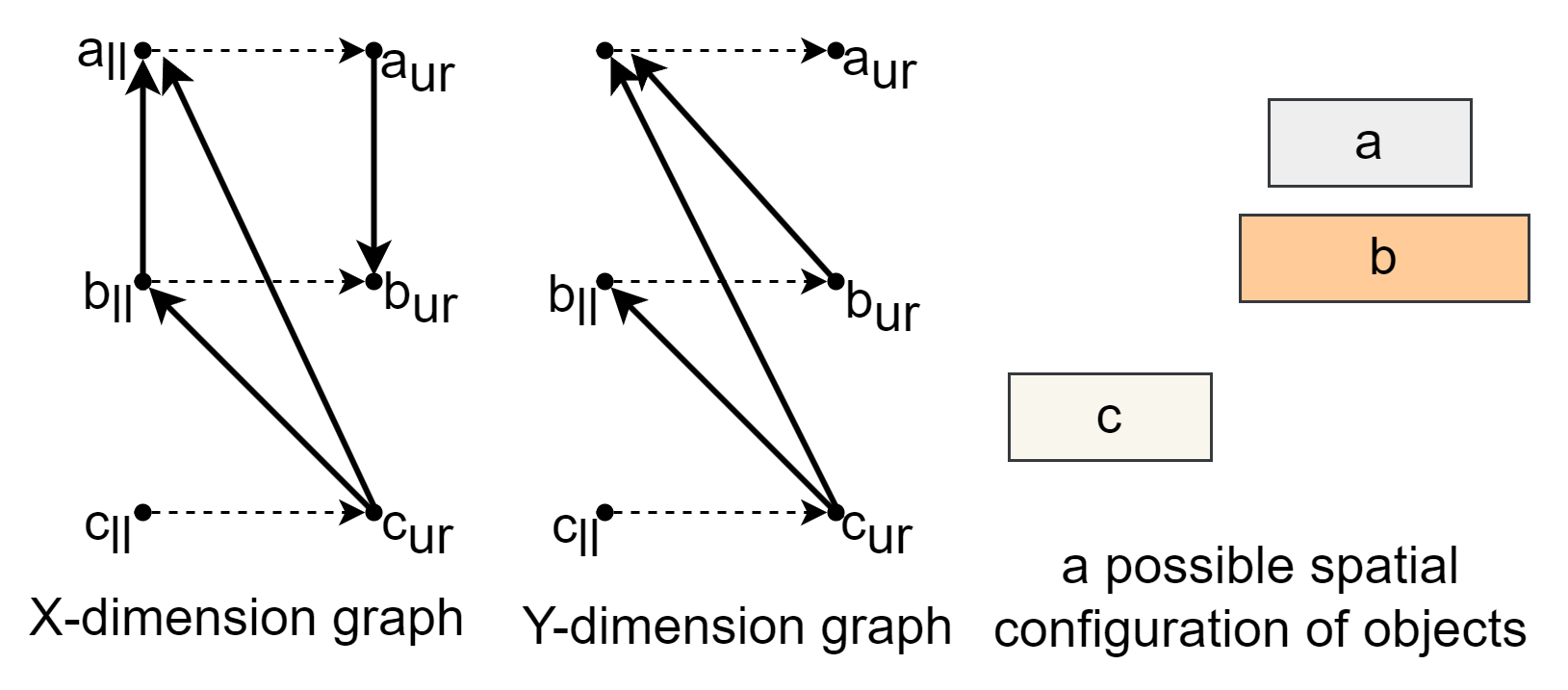}
    \caption{Example dimension graphs and a spatial configuration for the conjunctive constraint N(\textit{a,b})$\wedge$NE(\textit{b,c})$\wedge$SW(\textit{c,a}). The objects are represented by their MBR using lower left (ll) and upper right (ur) points. Continuous arrows represent $\leq$ constraints, and dotted arrows represent intrinsic constraints between ll and ur of an MBR \cite{dimensionGraphApproach}.}
    \label{example-dimension-graphs}
\end{figure}

At this stage of task generation, we have a set of spatial constraints among objects in the 2D Euclidean space. These spatial constraints are denoted as point connections using the above-mentioned 5-point representation of the objects. Now, the task at hand is to examine the consistency of these Euclidean spatial constraints among points. The purpose of consistency checking is to ensure the existence of a plausible spatial configuration that satisfies the constraints. To tackle this issue, we use an approach based on the dimension graph representation for managing the spatial constraints between objects \cite{dimensionGraphApproach}. In this technique, the spatial constraints are projected onto the X and Y dimensions, and the constraints that need to be met for each dimension are stored in separate graphs. From this technique, the problem of checking constraint consistency is reformulated as a graph cycle detection problem on the dimension graph. The constraints are considered consistent if both the X and Y dimension graphs are free of cycles. Figure \ref{example-dimension-graphs} illustrates the dimension graphs of a sample conjunctive constraint. It is noteworthy that the illustration in this example employs the MBR of the objects for the sake of simplicity, whereas in this study, a more complex 5-point representation is used.

During the construction of dimension graphs, constraints in a general format are converted to the Disjunctive Normal Form (DNF), which is a disjunction of conjunctions without any disjunction within a conjunction. For instance, a constraint graph with N(\textit{a,b})$\vee$NE(\textit{a,b}) and NE(\textit{b,c}) and SW(\textit{c,a}) can be transformed to (N(\textit{a,b})$\wedge$NE(\textit{b,c})$\wedge$SW(\textit{c,a}))$\vee$(NE(\textit{a,b})$\wedge$NE(\textit{b,c})$\wedge$SW(\textit{c,a})) in DNF format. Consequently, for a given constraint graph, there exists a pool of dimension graph pairs (for X and Y dimensions) that can be generated from all its conjunctions. The pool size is equivalent to the number of conjunctions in the DNF. In the generation process, a random pair of dimension graphs is selected, and its cycles are checked. If there are no cycles, the generation proceeds to the next steps; otherwise, another pair is chosen and tested. In case no pair in the pool without cycles (i.e., all possible combinations of constraints are inconsistent) is found, the scenario defined is considered unfeasible to accomplish in the 2D Euclidean space. Once a consistent set of constraints is determined, the forward checking technique is employed to solve them, and a set of possible values for the points (i.e., object positions) is obtained.

\subsection{Satisfying Constraints through Simulation}

Although qualitative methods can provide a useful starting point to narrow down the extensive and unbounded generative space, they may not be sufficient to fully address the challenges posed by physical environments. The physical environments present a multitude of challenges that are difficult to overcome through purely qualitative methods. Interactions between objects in these environments are complex and even slight variations in these interactions can result in significant changes in the overall outcome. As such, researchers have relied on simulation-based techniques when generating physics-based tasks \cite{stephenson2017generating, deceptiveGenerator}.

In this step of the generation process, the destroy interaction and the constraints of liesOnPath and pathObstructed are satisfied by simulating the physics engine of the game. The simulation begins by setting up the objects in the game space according to their predetermined positions from the previous steps. Then, the behaviours of objects are observed for constraint satisfaction by executing potential solution actions (i.e., shooting the bird at the object that initiates the interaction sequence). However, due to the continuous action space in Angry Birds (where shooting angles can be chosen from a continuous range), there exist numerous actions with slight variations. Therefore, the possible solution space is discretized by taking into account the target points and the stretch of the slingshot when shooting the bird. Specifically, only certain points of interest on the target object (such as the ll, ul, and ur) are selected, and it is assumed that the bird is always shot with the full stretch of the slingshot, resulting in only two possible trajectories to reach a target point. Shooting with the full stretch reflects the current capability of Angry Birds playing agents \cite{phyq}.

The object destruction model in the game is based on the health points which diminish through collisions and is mainly reliant on the materials of the colliding objects and the relative velocities they attain during the collision. In order to ensure that the destroy interaction is fulfilled, the health points of the objects of interest are closely monitored at the relevant juncture of the interaction sequence.

To ensure that the liesOnPath and pathObstructed constraints are satisfied, precise consideration of the moving paths of the objects is required. 
The simulation results are used to observe the paths of the objects associated with these constraints. To satisfy liesOnPath(\textit{a})(\textit{b}), the position of the object \textit{a} is adjusted to the location that intersects the most paths of \textit{b} for different actions from the solution actions. After repositioning, the new object configuration is verified by checking whether it still satisfies all the spatial constraints considered in the previous steps. If not, the process is repeated by repositioning object \textit{a} to the next position that cuts the most paths, and so on. Once the liesOnPath constraints are satisfied, the pathObstructed constraints are handled. The pathObstructed(\textit{a})(\textit{b}) constraint is satisfied by blocking the paths of \textit{a} that lead to \textit{b}, by adding platform objects as obstacles. The simulation results and QSRs between the objects are used to determine the paths of the object \textit{a} that lead to \textit{b}. It is ensured that the added obstacles do not interfere with the defined interaction sequence of the task.

If any of the simulation satisfactions fail during the generation process, the approach retries with a different dimension graph pair from the pool of dimension graphs, which is associated with a different conjunctive constraint in the DNF. This process is repeated until a feasible dimension graph pair is found. If the entire pool of dimension graphs is exhausted without finding a feasible solution, it is concluded that the defined scenario is infeasible to achieve.

\subsection{Incorporating Distractions}
\label{distractions-and-solvability-validation}
In this work, the main purpose of generating these tasks is to use them to assess the physical reasoning capabilities of AI agents. But when solving the tasks, agents may rely on spurious patterns instead of reasoning about the underlying physics. To prevent the exploitation of such patterns, we introduce a random number of distracting objects at arbitrary positions within the generated tasks. This approach aims to reduce the likelihood of agents exploiting specific patterns, such as `shoot the bird at the path unobstructed block', and encourages agents to engage in genuine physical reasoning.

\subsection{Ensuring the Intended Solvability}
In the final step, the solvability of the task is verified by executing the intended solution, which entails shooting the bird at the target object that initiates the interaction sequence. Upon successful completion, the output of the generator is a game level that possesses a definite solution consistent with the defined causal sequence of physical interactions.

\section{Results and Evaluations}

\begin{figure*}[!ht]
    \centering
    
    \begin{subfigure}{0.245\textwidth}
        \includegraphics[width=\textwidth]{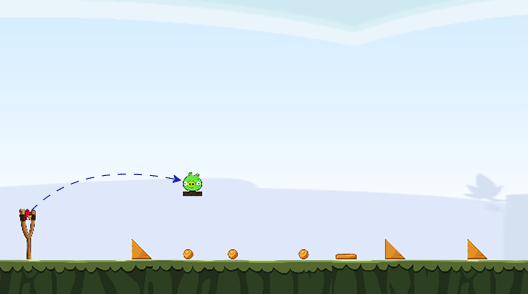}
        \caption{1. SF}
        \label{example-taks-of-16-scenarios:subfig1}
    \end{subfigure}
    \hfill
    \begin{subfigure}{0.245\textwidth}
        \includegraphics[width=\textwidth]{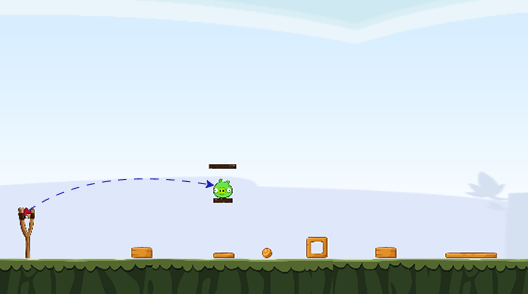}
        \caption{2. SFTB}
        \label{example-taks-of-16-scenarios:subfig2}
    \end{subfigure}
    \begin{subfigure}{0.245\textwidth}
        \includegraphics[width=\textwidth]{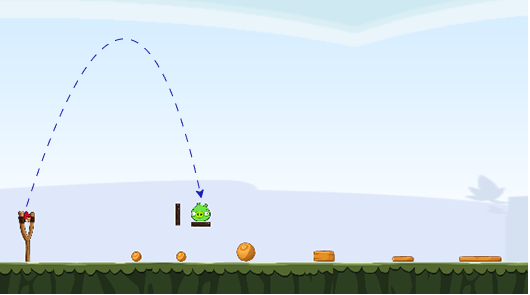}
        \caption{3. SFLB}
        \label{example-taks-of-16-scenarios:subfig3}
    \end{subfigure}
    \hfill
    \begin{subfigure}{0.245\textwidth}
        \includegraphics[width=\textwidth]{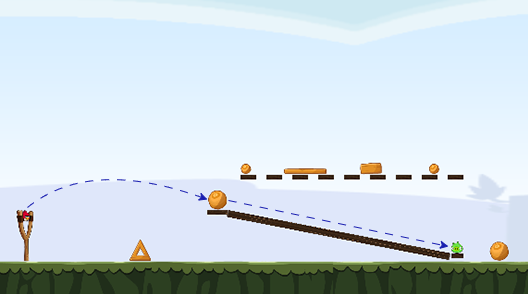}
        \caption{4. R}
        \label{example-taks-of-16-scenarios:subfig4}
    \end{subfigure}

    \begin{subfigure}{0.245\textwidth}
        \includegraphics[width=\textwidth]{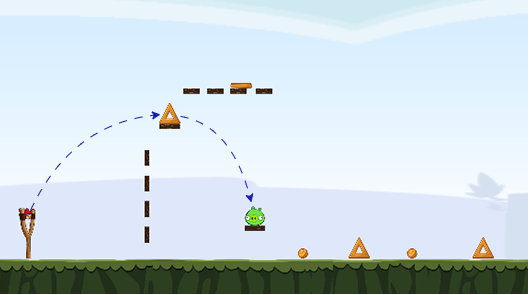}
        \caption{5. F}
        \label{example-taks-of-16-scenarios:subfig5}
    \end{subfigure}
    \hfill
    \begin{subfigure}{0.245\textwidth}
        \includegraphics[width=\textwidth]{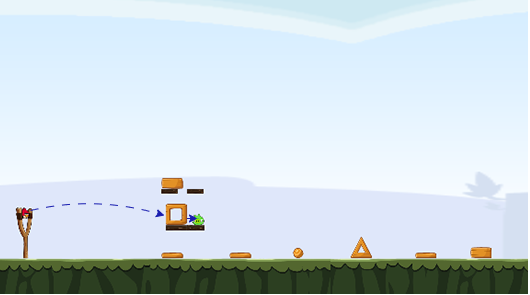}
        \caption{6. S}
        \label{example-taks-of-16-scenarios:subfig6}
    \end{subfigure}
    \begin{subfigure}{0.245\textwidth}
        \includegraphics[width=\textwidth]{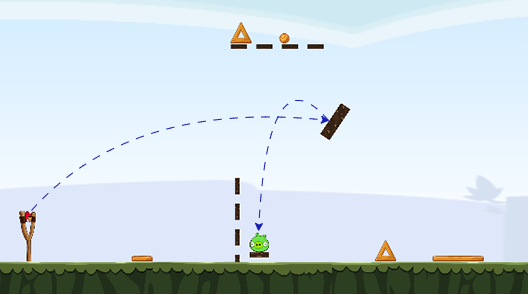}
        \caption{7. B}
        \label{example-taks-of-16-scenarios:subfig7}
    \end{subfigure}
    \hfill
    \begin{subfigure}{0.245\textwidth}
        \includegraphics[width=\textwidth]{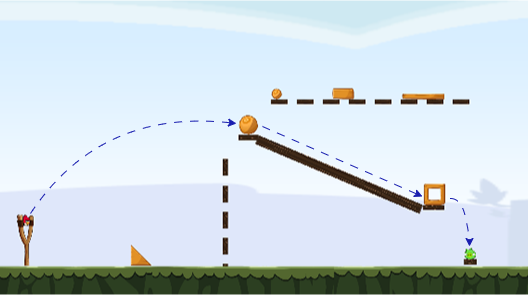}
        \caption{8. RF}
        \label{example-taks-of-16-scenarios:subfig8}
    \end{subfigure}

    \begin{subfigure}{0.245\textwidth}
        \includegraphics[width=\textwidth]{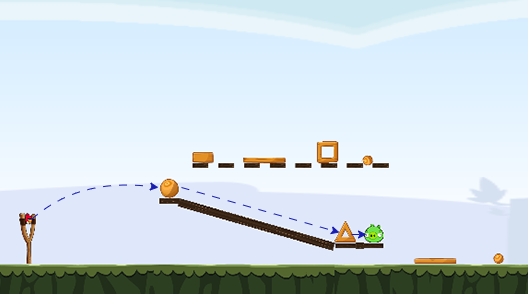}
        \caption{9. RS}
        \label{example-taks-of-16-scenarios:subfig9}
    \end{subfigure}
    \hfill
    \begin{subfigure}{0.245\textwidth}
        \includegraphics[width=\textwidth]{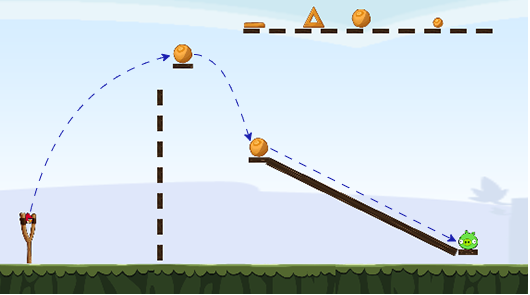}
        \caption{10. FR}
        \label{example-taks-of-16-scenarios:subfig10}
    \end{subfigure}
    \begin{subfigure}{0.245\textwidth}
        \includegraphics[width=\textwidth]{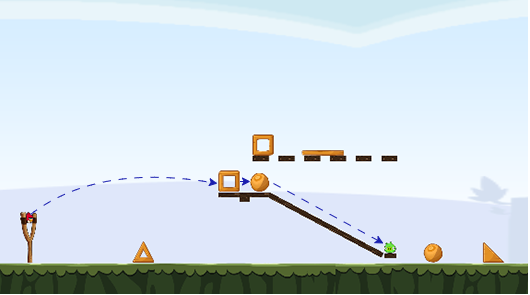}
        \caption{11. SR}
        \label{example-taks-of-16-scenarios:subfig11}
    \end{subfigure}
    \hfill
    \begin{subfigure}{0.245\textwidth}
        \includegraphics[width=\textwidth]{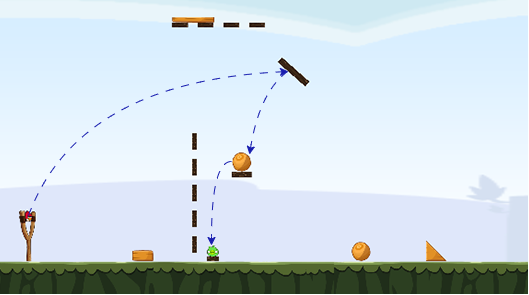}
        \caption{12. BF}
        \label{example-taks-of-16-scenarios:subfig12}
    \end{subfigure}

    \begin{subfigure}{0.245\textwidth}
        \includegraphics[width=\textwidth]{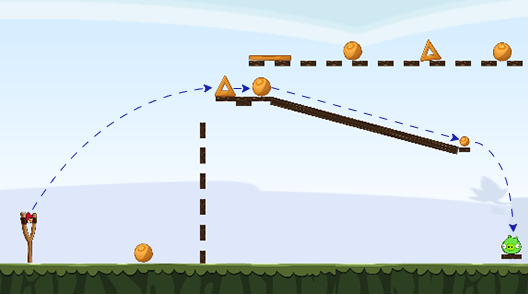}
        \caption{13. SRF}
        \label{example-taks-of-16-scenarios:subfig13}
    \end{subfigure}
    \hfill
    \begin{subfigure}{0.245\textwidth}
        \includegraphics[width=\textwidth]{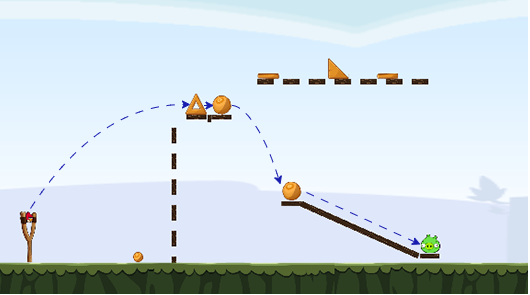}
        \caption{14. SFR}
        \label{example-taks-of-16-scenarios:subfig14}
    \end{subfigure}
    \begin{subfigure}{0.245\textwidth}
        \includegraphics[width=\textwidth]{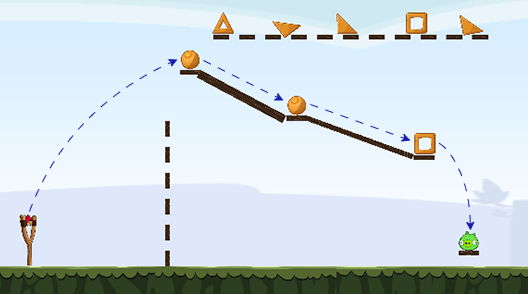}
        \caption{15. RRF}
        \label{example-taks-of-16-scenarios:subfig15}
    \end{subfigure}
    \hfill
    \begin{subfigure}{0.245\textwidth}
        \includegraphics[width=\textwidth]{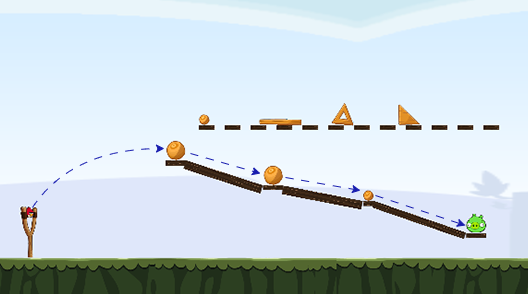}
        \caption{16. RRR}
        \label{example-taks-of-16-scenarios:subfig16}
    \end{subfigure}
    
    \caption{
    Generated tasks for the 16 example scenarios. Arrows show the object trajectories when the solution is executed.}
    \label{example-taks-of-16-scenarios}
\end{figure*}

    
    

In this section, we assess the performance of our proposed method. We begin by presenting Figure \ref{example-taks-of-16-scenarios}, which showcases 16 levels that were generated for the sample scenarios provided in Table \ref{example-scenario-definitions}. 

To evaluate the generated tasks, we conducted a series of analyses related to physical stability, solvability using intended solutions, and accidental solvability using unintended solutions. We generated a total of 480 tasks, with 30 tasks for each of the 16 sample scenarios. The evaluation results are presented as an aggregate, taking into account the number of physical interactions defined in each scenario. Specifically, scenarios 1 to 3 have two interactions, 4 to 7 have four interactions, 8 to 12 have six interactions, and 13 to 16 involve eight interactions. Furthermore, we analyzed the runtime of the generator to provide insights into the efficiency of our proposed method.

\subsection{Physical Stability}

\begin{table*}[h!]
  \small
  \centering
    {\fontsize{9}{11}\selectfont
    \begin{tabular}{lllll}
      \hline
      \multirow{2}{*}{Evaluation Metric} & \multicolumn{4}{c}{Physical Interaction Count in the Scenario} \\
       \cline{2-5}  
      & \multicolumn{1}{c}{Two} & \multicolumn{1}{c}{Four} & \multicolumn{1}{c}{Six} & \multicolumn{1}{c}{Eight} \\
      \hline
      Intended Solvability Rate & $1.00\pm0.00$ & $0.91\pm0.11$ & $0.87\pm0.10$ & $0.82\pm0.07$ \\
      \hline
      Accidental Solvability Rate & $0.12\pm0.05$ & $0.08\pm0.03$ & $0.07\pm0.01$ & $0.03\pm0.01$ \\
      \hline
      Generation Time (Seconds) & $1.83\pm0.10$ & $2.37\pm0.27$ & $3.16\pm0.86$ & $4.68\pm1.03$ \\
      \hline
    \end{tabular}}
  \caption{The results of the intended solvability, accidental solvability, and generation time evaluations (Mean$\pm$SD).}
  \label{evaluation-results}
\end{table*}

In physical environments, ensuring the physical stability of the tasks is crucial. Despite the absence of complex physical structures in the tasks generated through the proposed method, ensuring the physical stability of all the dynamic objects in a task is essential for overall stability. As discussed previously, when creating the layout constraint graph, the physical stability of dynamic objects is ensured by introducing supports. Additionally, during the last stage of generation, distracting objects are placed only in locations that can sufficiently support their stability. The generation process strictly enforces these stability checks, resulting in a 100\% physical stability rate for tasks across all scenarios.

\subsection{Intended Solvability}

In this evaluation, we assess the solvability of the generated tasks using the intended sequence of physical interactions defined in the scenario used to generate the task. To determine the solvability rate, the solution of each task (i.e., the shooting angle of the bird) was recorded during the task generation process and then executed using a playtest agent. Notably, the final step of the generation process that usually validates solvability was skipped in this experiment to evaluate how often the proposed methodology generates a solvable task. The solvability rate was calculated as the percentage of tasks that can be solved using their intended solutions. The results, shown in Table \ref{evaluation-results}, demonstrate that the solvability rate decreases as the number of interactions in the scenario increases. 
Specifically, the highest rate is 100\% for tasks with two interactions, while the lowest rate is 82\% for tasks with eight interactions.

\subsection{Accidental Solvability}

There is a potential for unintended interactions leading to solving the tasks in a continuous physical environment, which can be exploited by AI agents, thereby limiting the credibility of the evaluations conducted using those tasks. We propose the accidental solvability evaluation to assess the vulnerability of the generated tasks to unintended solutions. 
The evaluation involves the deployment of a heuristic agent, that systematically shoots at all dynamic blocks in the game level, excluding the target block for the solution. The heuristic of  shooting at different types of objects located at different positions is widely used by many Angry Birds agents \cite{2017AIBirdsComp}.

\begin{equation}
\label{accidential-solvability-equation}
\resizebox{0.7\linewidth}{!}{$
AS_i = \dfrac{1}{N_i}\sum_{n=1}^{N_i} \dfrac{1}{P_n} \sum_{p=1}^{P_{n}} (S_{np})
$}
\end{equation}


The accidental solvability rate ($AS_i$) of a scenario $i$ is calculated using Equation \ref{accidential-solvability-equation}, where $N_i$ is the total number of levels tested, $P_n$ is the total number of plays used to test the $n^{th}$ level, and $S_{np}$ is 1 if the $n^{th}$ level is solved by the $p^{th}$ strategy, or 0 otherwise. The $AS_i$ value ranges between 0 and 1, and a higher value indicates a higher vulnerability to unintended solutions. The results shown in Table \ref{evaluation-results} demonstrate a decreasing trend in accidental solvability rate as the interaction count of the scenario increases. Specifically, the highest rate is 12\% for tasks with two interactions, while the lowest is 3\% for tasks with eight interactions.


\subsection{Generation Time}


The task generator is implemented in C\# within the Unity game engine and integrated into the Angry Birds clone, Science Birds. The simulation-based components of the generator were executed by accelerating the physics engine by a factor of five. The experiments were performed on a Windows 10 desktop computer with an i9-9900KS CPU and 64GB RAM. The runtime of the generator was measured by recording the time taken to produce tasks for each sample scenario. The last row in Table \ref{evaluation-results} presents the results of the runtime analysis, indicating that the generation time increases linearly with the complexity of the scenario.

\section{Conclusion and Future Work}

This research presented a novel method for generating physical tasks using a systematic approach that leverages physical scenarios defined as a causal sequence of physical interactions between objects. We first introduced a grammar consisting of object, interaction, restriction, and layout grammars to support defining scenarios and object configuration in physical environments. The physical scenarios defined, adhering to this grammar, were used as input to the task generation process. In the task generation phase, the qualitative spatial relationships between objects were inferred, and a spatial constraint graph was constructed. For the narrowed-down generative space using spatial constraints, simulation-based constrain satisfactions were performed to determine the precise positions of the objects. We conducted evaluations on the generated tasks, focusing on physical stability, intended solvability, and accidental solvability. Additionally, we analyzed the task generation time. The results indicate that the proposed method is competent in generating tasks for a given causal sequence of physical interactions, achieving satisfactory performance across the evaluated metrics.

Moving forward, we foresee several potential paths of improvement for this research. One such direction would be to expand the current limitations of the process, allowing the generation of tasks that consider multiple interaction sequences with interdependent effects. Additionally, the proposed methodology can be directly applied to physics-based action domains like PHYRE, OGRE, and Virtual Tools, where the agent's action involves a one-time event, similar to shooting a bird in Angry Birds.
Furthermore, investigating the applicability of this approach in domains such as Animal-AI testbed \cite{crosby2020animal}, where the agent performs continuous interventions in the environment to solve tasks, holds promise for future exploration. Overall, this research opens up a new line of inquiry in the field of physics-based task generation, and its outcomes will be useful in developing AI with better physical reasoning capabilities.

\bibliography{aaai23}

\end{document}